\documentclass[11pt, letterpaper]{article}
\usepackage[table]{xcolor}
\usepackage{amsmath}
\usepackage{amssymb}
\usepackage{amsthm}
\usepackage{graphicx}
\usepackage{hyperref}
\usepackage{url}
\usepackage{algorithm}
\usepackage{algorithmic}
\usepackage{caption}
\usepackage{subcaption}
\usepackage{booktabs}
\usepackage{authblk}
\usepackage{appendix}
\usepackage[utf8]{inputenc}
\usepackage[T1]{fontenc}
\usepackage[english]{babel}
\usepackage{geometry}
\usepackage{xcolor}
\usepackage{fancyhdr}
\usepackage{microtype}
\usepackage{parskip}
\usepackage{tabularx}
\usepackage{enumitem}
\usepackage{multirow}
\usepackage{array}
\usepackage{booktabs}
\usepackage{ragged2e}
\usepackage{colortbl}
\usepackage{longtable}
\usepackage{adjustbox}
\hypersetup{
    colorlinks=true,
    linkcolor=blue,
    filecolor=magenta,      
    urlcolor=cyan,
    citecolor=red,
}

\definecolor{headercolor}{RGB}{0, 50, 100}
\title{Ophtha-LLaMA2: A Large Language Model for Ophthalmology}

\newcommand*\samethanks[1][\value{footnote}]{\footnotemark[#1]}
\author[1]{Huan Zhao\thanks{Co-first authors.}}
\author[2,3]{Qian Ling\samethanks}
\author[1,4]{Yi Pan\samethanks}
\author[5]{Tianyang Zhong\samethanks}

\author[2]{Jin-Yu Hu\thanks{Co-second authors.}}
\author[1]{Junjie Yao\samethanks}
\author[1]{Fengqian Xiao\samethanks}
\author[1]{Zhenxiang Xiao\samethanks}
\author[6]{Yutong Zhang\samethanks}
\author[2]{San-Hua Xu\samethanks}
\author[2,7]{Shi-Nan Wu\samethanks}
\author[2]{Min Kang\samethanks}
\author[4]{Zihao Wu\samethanks}
\author[4]{Zhengliang Liu\samethanks}
\author[1]{Xi Jiang\thanks{Co-corresponding authors. E-mails: xijiang@uestc.edu.cn, tliu@uga.edu, freebee99@163.com}}
\author[4]{Tianming Liu\samethanks}
\author[2,3]{Yi Shao\samethanks}

\affil[1]{The Clinical Hospital of Chengdu Brain Science Institute, MOE Key Laboratory for NeuroInformation, School of Life Science and Technology, University of Electronic Science and Technology of China, Chengdu 611731, China}
\affil[2]{Department of Ophthalmology, The First Affiliated Hospital of Nanchang University, Jiangxi Medical College, Nanchang, Jiangxi 330006, China}
\affil[3]{Department of Ophthalmology, Eye \& ENT Hospital of Fudan University, Shanghai 200030, China}
\affil[4]{School of Computing, University of Georgia, Athens 30602, USA}
\affil[5]{School of Automation, Northwestern Polytechnical University, Xi'an 710072, China}
\affil[6]{Institute of Medical Research, Northwestern Polytechnical University, Xi'an 710072, China}
\affil[7]{Eye Institute of Xiamen University, School of Medicine, Xiamen University, Xiamen, Fujian 361000, China}

\date{}

\begin{document}

\maketitle{}

\begin{abstract}
In recent years, pre-trained large language models (LLMs) have achieved tremendous success in the field of Natural Language Processing (NLP). Prior studies have primarily focused on general and generic domains, with relatively less research on specialized LLMs in the medical field. The specialization and high accuracy requirements for diagnosis in the medical field, as well as the challenges in collecting large-scale data, have constrained the application and development of LLMs in medical scenarios. In the field of ophthalmology, clinical diagnosis mainly relies on doctors' interpretation of reports and making diagnostic decisions. In order to take advantage of LLMs to provide decision support for doctors, we collected three modalities of ophthalmic report data and fine-tuned the LLaMA2 model, successfully constructing an LLM termed the “Ophtha-LLaMA2” specifically tailored for ophthalmic disease diagnosis. Inference test results show that even with a smaller fine-tuning dataset, Ophtha-LLaMA2 performs significantly better in ophthalmic diagnosis compared to other LLMs. It demonstrates that the Ophtha-LLaMA2 exhibits satisfying accuracy and efficiency in ophthalmic disease diagnosis, making it a valuable tool for ophthalmologists to provide improved diagnostic support for patients. This research provides a useful reference for the application of LLMs in the field of ophthalmology, while showcasing the immense potential and prospects in this domain.
\end{abstract}

\section{Introduction}

With the rapid advancement of artificial intelligence, large language models (LLMs) have revolutionized Natural Language Processing (NLP), enabling computers to interact with textual and spoken language in a manner similar to humans. One notable example is the Generative Pretrained Transformer 3 (GPT-3), which has been extensively pre-trained on massive amounts of text data, allowing it to analyze and generate text in various healthcare domains.
These LLMs have been found wide applications in the healthcare field through pre-training on a vast amount of textual data and abstract analysis of texts, such as predicting postoperative hospitalization time and generating electronic health records\cite{danilov2022length,wang2022leveraging}. This indicates the potential applications of LLMs in clinical\cite{holmes2023benchmarking,liu2023artificial,liu2023evaluating,li2023artificial,holmes2023evaluating}, educational\cite{wu2023matching,latif2023artificial,liu2023context}, and research settings\cite{ouyang2022training, thirunavukarasu2023trialling}.

However, the application of LLMs in biomedicine has not been fully explored so far, and general LLMs lack pre-training on clinical medical expertise. Given the continuous research and innovation in the field of medical science, it is crucial to train and fine-tune models with up-to-date content to reduce the inaccuracies \cite{thirunavukarasu2023large} in LLM outputs and avoid the occurrence of hallucinations. Furthermore, there are limitations as LLMs like ChatGPT have inherently limited knowledge base and cannot access the internet in real-time while answering user queries. Some latest LLMs, such as BlenderBot3 \cite{shuster2022blenderbot} and Sparrow \cite{glaese2022improving}, can access the internet simultaneously while generating responses, providing additional sources of information. However, clinical case information systems typically contain personal health identifiers, including detailed patient information such as name, gender, age, as well as past medical history and disease diagnosis-related information \cite{moqurrab2021accurate}. While this information is essential for clinical diagnosis, it is important to address privacy protection concerns related to patient data associated with these clinical entities when it comes to LLMs.

Ophthalmology is a discipline that relies on various diagnostic techniques, including optical coherence tomography (OCT), ocular surface analyzer (OSA), and color fundus photography (CFP), and so on. Ophthalmologists typically rely on these imaging examination reports to make clinical diagnoses. However, there is currently a lack of a multimodal, self-service diagnostic decision system based on LLM construction. Therefore, in this study, our aim is to develop an LLM specifically for ophthalmology, called Ophtha-LLaMA2, with a clear and concise style of communication similar to that of ophthalmologists, based on the open-source LLMs like LLaMA2, and deploy them locally. With Ophtha-LLaMA2, we can further explore the potential of developing a text-visual aligned self-service diagnostic decision system, providing excellent performance and results for ophthalmology as well as other medical fields.

\section{Related Work}

\subsection{Evaluating LLMs}
LLMs have been proven to achieve superior performance in a variety of tasks and have thus gained popularity in many fields. These LLMs are characterized by their few-shot and zero-shot learning capabilities, which sets them apart from traditional pre-training and fine-tuning methods like BERT\cite{devlin2018bert}, Palm\cite{chowdhery2022palm} and GPT series\cite{floridi2020gpt,openai2023gpt4}.

Researchers typically assess the proficiency of various LLMs by subjecting them to benchmark datasets spanning diverse domains, including literature, chemistry, and biology, and subsequently evaluate their effectiveness using conventional metrics, such as accuracy, recall, and F1 score. A recent study \cite{yang2023dawn} by Microsoft delves into GPT-4V's \cite{openai2023gpt4v} qualitative capabilities by assessing its performance across 10 essential tasks, including open-world visual understanding, visual description, multimodal knowledge, offering a valuable array of techniques to enhance the effective use of large multimodal models. Another study conducted by Microsoft \cite{nori2023capabilities} showed that GPT-4 significantly outperforms the United States Medical Licensing Examination (USMLE), indicating its potential for success in the medical field.

To save computational resources and time, it is essential to select one suitable LLM from the numerous high-performing ones for fine-tuning to obtain a specialized LLM in ophthalmology. Therefore, evaluating these models and considering the ethical use and effectiveness of LLMs are crucial. This research represents an initial effort to comprehensively investigate LLMs in the field of ophthalmology, laying the foundation for future studies that primarily focus on evaluating the effectiveness of LLMs in complex professional medical practices.

\subsection{Applications of LLMs in Medical Field}
LLMs have achieved significant success in the general domain, and there have been advancements in open-source fine-tuning of these LLMs. Therefore, there is a growing interest in applying LLMs in the medical field. Fine-tuned LLMs demonstrate considerable potential in the biomedical domain, as they can perform various tasks, including interacting with patients, providing diagnoses and treatment recommendations, and explaining medical knowledge.

Prior researches have shown that LLMs can be used to recommend medication \cite{harskamp2023performance}, determine the cause of the disease, or determine imaging services based on a patient's clinical presentation\cite{rao2023evaluating}. In addition, LLMs can predict the treatment effect and prognosis status based on the patient's medical history and similar cases, which can help physicians to select appropriate treatment options and make decisions. Specifically, LLMs can be applied in four major areas: clinical literature, clinical decision support, knowledge-based medical information retrieval and generation, and medical research. For clinical doctors and professionals, completing clinical documents, including but not limited to medical records, test reports, and discharge letters, accounts for a significant portion of their workload. Previous studies have suggested that LLMs trained on large volumes of text data can be applied in the healthcare field. These models can assist in documenting patient information and symptoms, help with writing discharge letters, automatically generate clinical histories and examination reports, thereby reducing the writing burden on doctors.

In addition, knowledge-based healthcare information retrieval and creation is another area where LLMs can play an important role. For example, LLMs are packaged as an application to answer health-related questions posed by patients\cite{liu2023summary,nov2023putting,shi2023mededit}. By integrating the questions asked by the patient with the patient's past medical records, LLMs can provide personalized information and provide information on anything that may be relevant to the questions asked by the patient, such as medication usage. It has also been proposed that LLMs can be used in medical research\cite{liu2023summary,wang2023prompt,zhang2023biomedgpt,liu2023pharmacygpt,cai2023exploring,zhong2023chatabl,wang2023review} and public health\cite{dai2023ad,guan2023cohortgpt,zhong2023chatradio}.LLMs can further help protect patient privacy\cite{liu2023deid}, enhance medical text data\cite{dai2023auggpt}, and so on.

\section{Method}
Ophtha-LLaMA2 is trained on a private ophthalmology dataset using fine-tuning to generate ophthalmology impressions from imaging examination report findings. Both engineering and clinical indicators have confirmed that Ophtha-LLaMA2 is highly reliable, concise, and clinically useful in generating these impressions, providing important guidance.

\subsection{Ophthalmology Dataset}

\subsubsection{Introduction to Ophthalmic Data}

Our training and testing data covers various types of data in the field of ophthalmology, including data on the Meibomian glands, fundus, and retina, acquired by OCT, OSA, and CFP, respectively.

OSAs are specifically designed to detect and evaluate the function and structure of the meibomian glands. Using this device, ophthalmologists can observe and assess the secretion of the meibomian glands, evaluate the morphology of gland orifices, assess gland loss, and identify duct obstructions. These observations provide vital guidance for clinical decision making.

CFP is an important technique used to observe and record the condition of the fundus. By employing this device for fundus imaging, ophthalmologists can directly observe structures such as blood vessels, the retina, optic disc, and macula. This enables timely detection and diagnosis of various retinal diseases, including diabetic retinopathy and glaucoma.

OCT is a non-invasive imaging technology that provides high-resolution cross-sectional imaging of ocular tissue. This technique generates images by measuring light reflections and interference. OCT aids ophthalmologists in observing and evaluating abnormalities in structures such as the retina, vitreous, and optic nerve. It is especially useful for early diagnosis and monitoring of diseases such as macular degeneration and retinal detachment.

\begin{figure}[htbp]
    \centering
    \includegraphics[width=0.4\linewidth]{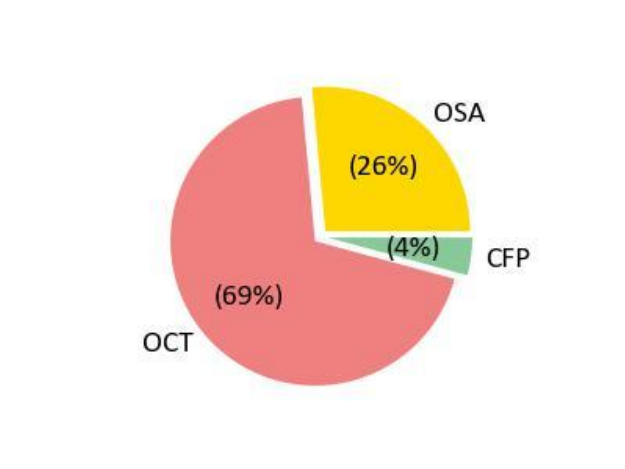}
    \caption{The proportion of different modalities of collected samples}
    \label{fig:samples_proportion}
\end{figure}

The aforementioned three imaging devices are the most common pieces of equipment used in the field of ophthalmology, and these three imaging modalities play a crucial role in the diagnosis and treatment of ocular diseases. Specifically, we collected 19,539 cases involving OSA, 51,146 cases of CFP, and 3,105 cases of OCT. Figure \ref{fig:samples_proportion} displays the distribution of these cases. Although these cases are from different patients, there may be some duplicates or overlapping information regarding the related conditions. Therefore, data cleansing is necessary when utilizing the data in practice.

\subsubsection{Data Preprocessing}
In the overall framework of Ophtha-LLaMA2, the quality of training data plays a crucial role in enhancing system performance. Data preprocessing involves the meticulous organization and cleansing of multimodal and multi-disease ophthalmic data as described in this paper, aiming to ensure data accuracy and consistency.

\textbf{(1) Data Cleaning}

To collect the raw data, we collected examination report descriptions and doctor diagnoses, which are key clinical records, generated by three imaging devices: OSA\cite{xu2022altered}, CFP\cite{liao2019altered}, and OCT\cite{shao2018visualization,ye2018retinal}. Specifically, during the data collection process, each patient's examination report for each imaging modality was treated as a sample. These examination reports consisted of the patient's personal information (e.g., name, gender, age), imaging results from the devices, automated disease descriptions based on the imaging results, as well as detailed information regarding the medical examination status and diagnosis made by the doctor based on the examination reports. We extracted the examination report descriptions and corresponding doctor diagnosis information from each sample, ensuring that privacy information was handled confidentially. Furthermore, we excluded data with possible misdiagnosis or requiring further examination for a diagnosis to maintain the validity of the data quality. 

\begin{figure}[!htbp]
    \centering
    \includegraphics[width=0.4\linewidth]{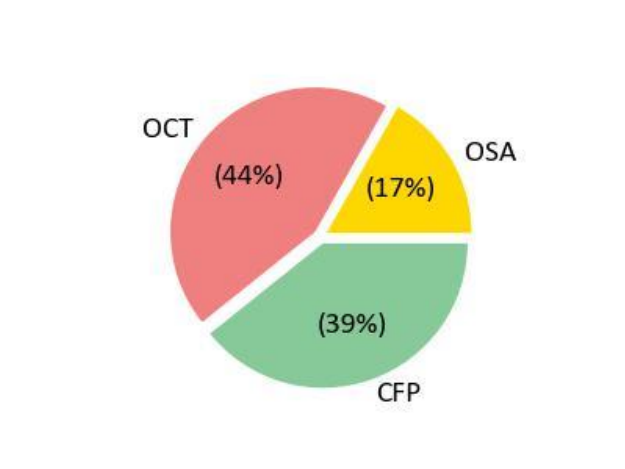}
    \caption{The proportion of different modalities of data in the dataset}
    \label{fig:data_proportion}
\end{figure}

\textbf{(2) Dataset Partitioning}

For the samples obtained from the above sampling, there are cases where different patients have the same symptoms and diseases. By removing samples of duplicate and unintentional items, we ultimately collected a total of 7065 data samples, as shown in Figure \ref{fig:data_proportion}, including 1189 samples of meibomian gland data, 2773 samples of fundus data, and 3103 samples of retinal data. It is worth noting that although the data volume is in the thousands, it represents information from over one hundred thousand cases. Finally, we divided these data into a training set and a test set in a ratio of 6:4.

\subsection{Framework Overview}

\begin{figure}[!htbp]
    \centering
    \includegraphics[width=1\linewidth]{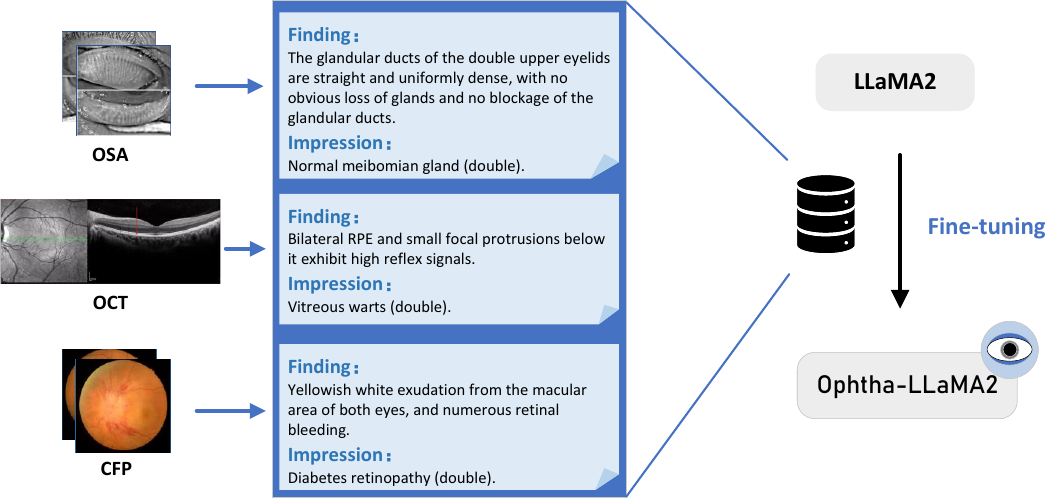}
    \caption{Overall framework}
    \label{fig:main1}
\end{figure}

For basic LLMs are trained on a wide range of general data, they have acquired strong reasoning abilities and knowledge reserves. However, when faced with more specialized problems, especially in the field of medicine, they often struggle to grasp the correct answers. To address this issue, an increasing number of researchers are fine-tuning LLMs using domain-specific data, making them experts in that particular field. In our study, as the overall framework shown in Figure \ref{fig:main1}, we fine-tuned the LLaMA2 model using ophthalmic data obtained from three types of ophthalmic examination instruments, transforming it into an expert in the field of ophthalmology, assisting ophthalmologists in disease screening and diagnosis.

\subsection{Fine-tuning Ophtha-LLaMA2}

\subsubsection{Architecture of the LLaMA2 Model}
LLMs represent a groundbreaking technological breakthrough in the field of deep learning. The open-source LLaMA model by Meta has garnered widespread attention. Recently, LLaMA2 has been shown to achieve remarkable performance in the areas of inference, encoding, capability, and knowledge testing, surpassing many other open-source language models.

The model structure of LLaMA2 \cite{touvron2023llama} is similar to the standard transformer decoder, as it consists of multiple transformer blocks. However, it also has some differences compared to the traditional Transformer Decoder:

\textbf{(1) Preceding RMSNorm layer} 

LLaMA2 adds a root mean square normalization (RMSNorm) layer before each transformer block to normalize the input features. This helps improve the model's adaptability to features of different scales.

\textbf{(2) RoPE positional encoding}

Before performing the multiplication operation between queries (Q) and keys (K), LLaMA2 uses relative positional encoding (RoPE) to encode the positions of the input sequence. Compared to traditional absolute positional encoding, RoPE better captures the relative positional relationships between different positions, thereby improving the model's performance on long sequences.

\textbf{(3) K V Cache and GQA} 

LLaMA2 introduces the K V Cache mechanism to store previously computed keys (K) and values (V), avoiding redundant computations, and reducing the computational load. Additionally, LLaMA2 uses group query attention (GQA), which partitions queries (Q) into different groups, to enhance the model's computational efficiency.

\textbf{(4) FeedForward Layer} 

LLaMA2 uses the gated linear units (GLU) structure in the FeedForward layer of each transformer block, improving the model's representation capability by adding gating mechanisms. The GLU structure learns gating weights to better capture the nonlinear relationships in the input features.

\begin{figure}[htbp]
    \centering
    \includegraphics[width=0.7\linewidth]{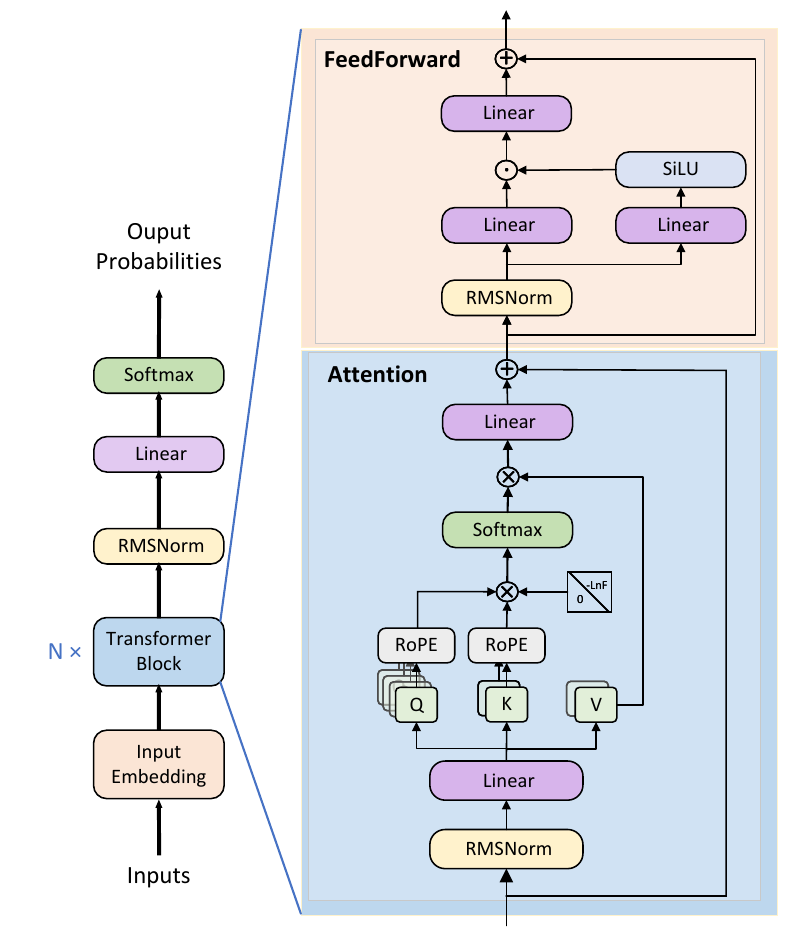}
    \caption{Architecture of LLaMA 2}
    \label{fig:main2}
\end{figure}

These differences above allow LLaMA2 to perform better and more efficiently in NLP tasks. With increased maximum input length and an expanded corpus, LLaMA2 has greater expressive power and reasoning ability compared to LLaMA. In summary, the improvements in LLaMA2 make it more suitable for handling complex NLP tasks.The overall network architecture of LLaMA 2 is shown in Figure \ref{fig:main2}.

\subsubsection{Quantization and Fine-Tuning}

\begin{figure}[!htbp]
    \centering
    \includegraphics[width=0.8\linewidth]{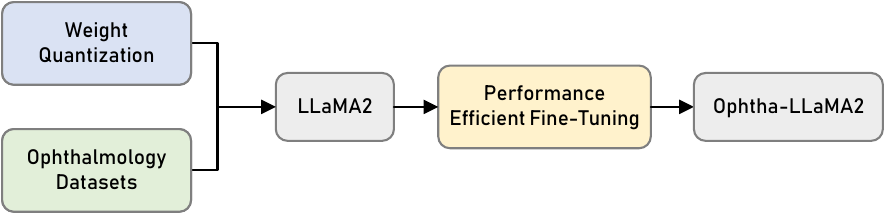}
    \caption{Fine-Tuning Process of LLaMA2}
    \label{fig:main3}
\end{figure}

As shown in Figure \ref{fig:main3}, the process of fine-tuning Ophtha-LLaMA2 based on LLaMA2 can be divided into three stages as follows:

\textbf{(1) Model Quantization and Data Preparation}

With the help of transformers' model quantization library, the BitsAndBytesConfig interface was used for model quantization. The purpose of BitsAndBytes is to reduce the floating-point precision of model weights through a quantization process from high precision to low precision. In this process, we convert the model weights to the int4 format through quantization layers and store them on the GPU. The core computations are performed on CUDA, which reduces memory consumption and improves efficiency. This allows us to fine-tune larger models on consumer-grade GPUs. A detailed explanation of the data preparation process is provided in section 3.1.

\textbf{(2) Pre-trained Model}

Before the training stage, we conducted a thorough analysis of well-known LLMs from a medical perspective. We then built a pool of LLMs to select the most suitable one for our purposes. Our evaluation primarily considered six key aspects: domain adaptability, compatibility with medical standards, bilingual capability, open source availability, parameter efficiency, and cost and licensing considerations.We chose several quantifiable evaluation metrics to demonstrate, such as the time required for fine-tuning and inference, and the similarity of inference results to ground truth.
In this experiment, we tested several open-source LLMs, including Baichuan, Bailing, ChatGLM2, InternLM, Ziya, and MOSS. However, as shown in Table \ref{table1Similarity} and Table \ref{table2Count}, LLaMA2 outperformed all of these models, showcasing superior performance. A detailed introduction to the LLaMA2 model is provided in section 3.3.1.

\textbf{(3) PEFT}

Based on the consideration of our limited dataset and to reduce training costs and mitigate the potential risk of catastrophic forgetting, we chose the Progressive Layer Freezing and Fine-tuning (PEFT) method to fine-tune LLaMA2. This method is currently one of the mainstream approaches for fine-tuning LLMs. PEFT selectively fine-tunes a small number of additional model parameters, significantly reducing computational and storage costs, allowing efficient adaptation of the pre-trained LLMs to various downstream application domains.

Specifically, we use low-rank adaptation (LoRA) \cite{hu2021lora} fine-tuning method to fine-tune the large language model. LoRA involves freezing the pre-training weights and creating low-rank versions of the attention matrices for the query and value layers. These low-rank matrices have significantly fewer parameters than the original model, allowing for fine-tuning with less GPU memory usage.

\begin{figure}[htbp]
    \centering
    \includegraphics[width=0.3\linewidth]{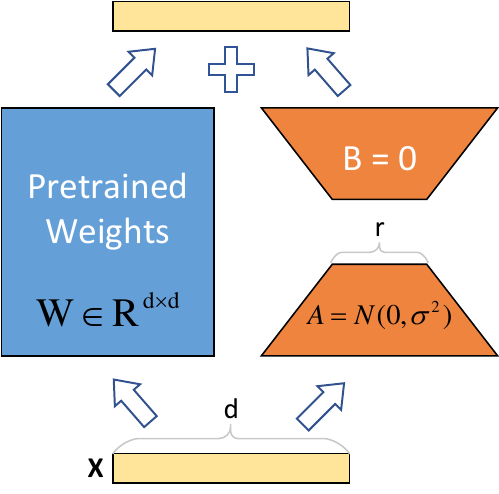}
    \caption{The Fine-Tuning Strategy for LoRA}
    \label{fig:main4}
\end{figure}

The principle of LoRA, as shown in Figure \ref{fig:main4}, can be divided into several main steps:
\begin{itemize}
  \item First, to adapt to a specific downstream task, we train a specific model to transform Y=WX  into Y=(W+$\Delta${W})X, where $\Delta${W} represents the results we want to fine-tune.
  \item Next, we perform low-rank decomposition on $\Delta${W}, represented as $\Delta${W}=AB (where $\Delta${W} is m*n-dimensional, A is m*r-dimensional, and B is r*n-dimensional, with r being the low-rank dimension assumed earlier).
  \item Then, by training A and B using specific training data, we obtain $\Delta${W}, which can be directly added to W during the inference process without any additional cost.
  \item Furthermore, if the LoRA needs to be switched to adapt different scenarios, the process can be simply achieved by performing matrix addition: (W + $\Delta${W}) - $\Delta${W} + $\Delta${W}'.
\end{itemize}

\subsection{Inference Ophtha-LLaMA2}

\subsubsection{Prompt Generation}
With the meticulously processed dataset proposed in Section 3.1, our objective is to generate complete prompts to be used by LLMs for generating effective inferences, contributing to both the training and evaluation phases. In addition to the corpus containing preprocessed multimodal ophthalmology data, we tailored LLMs instruction templates specifically for this study. These templates, designed with the input of domain experts, provide refined guidance to LLMs, enabling them to adapt to the ophthalmology domain and generate meaningful domain-specific results.The instruction template we used is shown in Figure \ref{fig:Results}.

\subsubsection{Text Summarization }
 
 We utilized the text summarization capability of LLMs to generate the diagnostic section of medical reports. By inputting the patient's clinical information and examination results, Ophtha-LLaMA2 is able to automatically extract key information and generate accurate and concise diagnostic summaries. This automated diagnostic generation process would help save time and effort for healthcare professionals, improving the efficiency and consistency of reports. Moreover, as further research progresses, the model can incorporate the latest medical knowledge and clinical guidelines to generate the report, ensuring the accuracy and reliability of the diagnostic results. Applying the text summarization capability of LLMs to the diagnostic generation of medical reports holds significant value in healthcare practice.

\subsection{Experimental Setting}

The hyperparameter settings used during the fine-tuning process using the lora method were shown in Table \ref{table4Fine-tune}:

    \begin{table*}[!htbp]
	    \centering
	    \caption{Fine-tune Hyperparameter Settings}
	    \label{table4Fine-tune}
	    \resizebox{1.0\linewidth}{!}{
	        \begin{tabular}{cccccc}
                \toprule[1.2pt]
                \footnotesize{Learning Rate} & \footnotesize{Batch Size} & \footnotesize{Max Sequence Length} & \footnotesize{Gradient Accumulation Steps} & \footnotesize{LoRA r} & \footnotesize{LoRA $\alpha$} \\
                \hline
                \footnotesize{$1.41 \times 10^{-5}$} & \footnotesize4 & \footnotesize512 &\footnotesize 16 &\footnotesize 64 & \footnotesize16 \\
                \bottomrule[1.2pt]
	        \end{tabular}
        }
    \end{table*}

During the inference stage, the hyperparameter settings were shown in Table \ref{table3Inference}:

\begin{table*}[!htbp]
	\centering
	\caption{Inference Hyperparameter Settings}\label{table3Inference}
	\resizebox{0.7\linewidth}{!}{
        \begin{tabular}{ccccc}
        \toprule[1.2pt]
  \footnotesize{Temperature} & \footnotesize{Max New Tokens} & \footnotesize{Repetition Penalty} & \footnotesize{Top k} & \footnotesize{Top p}\\ \hline  
	\footnotesize 0.9 & \footnotesize 512 & \footnotesize 1.3 & \footnotesize 40 & \footnotesize 0.9  \\
        \bottomrule[1.2pt]
	\end{tabular}}
\end{table*}

\section{Performance Evaluation}

In order to confirm the effectiveness, applicability, and portability of Ophtha-LLaMA2, we conducted a comprehensive analysis that included comparing the diagnostic results of the model with those of doctors, using multiple indicators to evaluate the performance of the model. Although we utilized a limited amount of data for fine-tuning, our results nevertheless demonstrated that Ophtha-LLaMA2 achieves favorable outcomes in Ophthalmology diagnostics.

\subsection{Quantitative Assessment}
Engineering metrics primarily utilize the Recall-Oriented Understudy for Gisting Evaluation (ROUGE) metric \cite{lin2004rouge}, which is an evaluation measure used to assess the similarity between automatically generated summaries or excerpts and reference summaries. ROUGE is based on the concept of text similarity, and calculates the overlap between the generated summary and the reference summary.

ROUGE has various variants, including ROUGE-N, ROUGE-L. 

ROUGE-N measures the overlap of N-grams. In the formula, the denominator calculates the number of n-grams in the report's diagnosis, while the numerator calculates the number of common n-grams between the doctor's diagnosis and the model's diagnosis.

\begin{equation}
ROUGE-N=\frac{\sum_{S\in ReferenceSummaries}\sum_{gram_{n}\in S}{Count_{match}(gram_{n})}}{\sum_{S\in ReferenceSummaries}\sum_{gram_{n}\in S}{ Count(gram_{n})}}
\end{equation}

ROUGE-L is an overlap measure based on the longest common subsequence (LCS), which calculates the length and ratio of the longest common word sequence between model and ophthalmologist diagnostics. LCS captures information about the word order, but does not require exact matching, making it more flexible than n-gram.
\begin{equation}
R_{LCS}=\frac{LCS(X,Y)}{m}
\end{equation}
\begin{equation}
P_{LCS}=\frac{LCS(X,Y)}{n}
\end{equation}
\begin{equation}
F_{LCS}=\frac{(1+\beta)^{2}R_{LCS}P_{LCS}}{R_{LCS}+\beta^{2}P_{LCS}}
\end{equation}

using the ROUGE metrics, we are able to quantitatively evaluate the similarity between the diagnoses generated by Ophtha-LLaMA2 based on the description in the examination reports and the diagnoses made by ophthalmologists. This assessment is highly beneficial in confirming the performance of the model.

\begin{table*}[!htbp]
	\centering
	\caption{Similarity between Ophtha-LLaMA2 and physician diagnosis}\label{table1Similarity}
        \fontsize{8.5}{13.8}\selectfont
	\resizebox{0.8\linewidth}{!}{%
	\begin{tabular}{m{.2\textwidth}<{}*{3}{m{.11\textwidth}<{\centering}}}
		\toprule[1.3pt]
		\multicolumn{1}{l}{\multirow{2}{*}{Model}} & \multicolumn{3}{c}{Rouge Scores}\\ \cline{2-4} & R-1 $\uparrow$ & R-2 $\uparrow$ & R-L $\uparrow$ \\ 
        \midrule[1.2pt]

	LLaMA2-7B	& 0.0722& 0.0149& 0.0207\\
        LLaMA2-7B-ft	& 0.3152& 0.1816& 0.2830\\\hline
        LLaMA2-chat-7B & 0.0527 &	0.0074&	0.0311\\
        Ophtha-LLaMA2 & \textbf{0.4866}&	\textbf{0.3886}& \textbf{0.4514}\\\hline
        LLaMA2-Chinese-7B	& 0.0584& 0.0098& 0.0383\\
        LLaMA2-Chinese-7B-ft	& 0.3954& 0.2523& 0.3367\\\hline
	Baichuan-7B & 0.0935& 0.0163&	0.0310\\
	Bailing-7B & 0.0482& 0.0072& 0.0285\\
 	ChatGLM2-6B & 0.0777& 0.0146&	0.0652\\
	InternLM-chat-7B & 0.0784& 0.0152& 0.0592\\
        InternLM-7B & 0.0415& 0.0056& 0.0268\\
        MOSS & 0.0375& 0.0055& 0.0125\\
        Ziya-13B & 0.0939 & 0.0189 &0.0808\\
        \bottomrule[1.3pt]
	\end{tabular}}
\end{table*}

\subsection{Generalization Performance for Multi-Imaging Modalities}

As stated in Section 3.1, the ophthalmic data used in this study was collected from three different types of instruments: optical coherence tomography, meibography systems, and fundus cameras. These instruments present data in different formats, and the focus of eye disease examination varies. During the fine-tuning and inference process, we randomly shuffled the reports from these modalities but ensured that both the training and test sets contain a comparable proportion of reports from all three modalities. Our objective was to develop a comprehensive ophthalmic model that can accurately and quickly diagnose diseases based on examination reports from different instruments. It is crucial to avoid missed diagnoses, misdiagnoses, and overdiagnoses in all imaging modalities, aiming to provide better eye health management and treatment options for patients.

\begin{table*}[!htbp]
	\centering
	\caption{Running Time Count}\label{table2Count}
        \fontsize{8.5}{13.8}\selectfont
	\resizebox{0.8\linewidth}{!}{
        \begin{tabular}{m{3.2cm}<{}m{2.8cm}<{\centering}m{2.8cm}<{\centering}}   
        \toprule[1.3pt]
           \textbf{Model} & \textbf{Fine-Tuning Time (hour)} & \textbf{Reference Time (second)} \\ 
        \midrule[1.2pt]
	LLaMA2-7B	& 2.3 & 14\\
        LLaMA2-7B-ft	& 2.3 & 14\\
        LLaMA2-chat-7B & 2.4 &	15\\
        LLaMA2-chat-7B-ft & 2.4&	15\\
        LLaMA2-Chinese-7B	& 2.4 & 12\\
        LLaMA2-Chinese-7B-ft	& 2.4 & 12\\\hline
	Baichuan-7B & null& 13\\
	Bailing-7B & null& 12\\
 	ChatGLM2-6B & null& 8\\
	InternLM-chat-7B & null& 13\\
        InternLM-7B & null& 12\\
        Ziya-13B & null & 21\\
        \bottomrule[1.3pt]
	\end{tabular}}
\end{table*}

\subsection{Computing Resource Evaluation}
We performed timing tests on multiple LLMs at different stages, including fine-tuning and inference testing. We use one NVIDIA GeForce RTX 3090 graphics card for inference (as the moss model has a large parameter size, we use two 3090 graphics cards), and two NVIDIA GeForce RTX 3090 graphics cards for fine-tuning. Based on the fine-tuning timing tests, as shown in Table /ref{table2}, we concluded that our fine-tuning approach is both effective and efficient, requiring minimal time. Furthermore, the inference timing tests showed that Ophtha-LLaMA2 does not require excessive computational power during the inference process, making it a low-energy and high-efficiency system for diagnosing eye diseases. We expect this model to become an efficient ophthalmic disease diagnostic system in the future.

\subsection{Result Analysis}

\begin{figure}[htbp]
    \centering
    \includegraphics[width=1.1\linewidth]{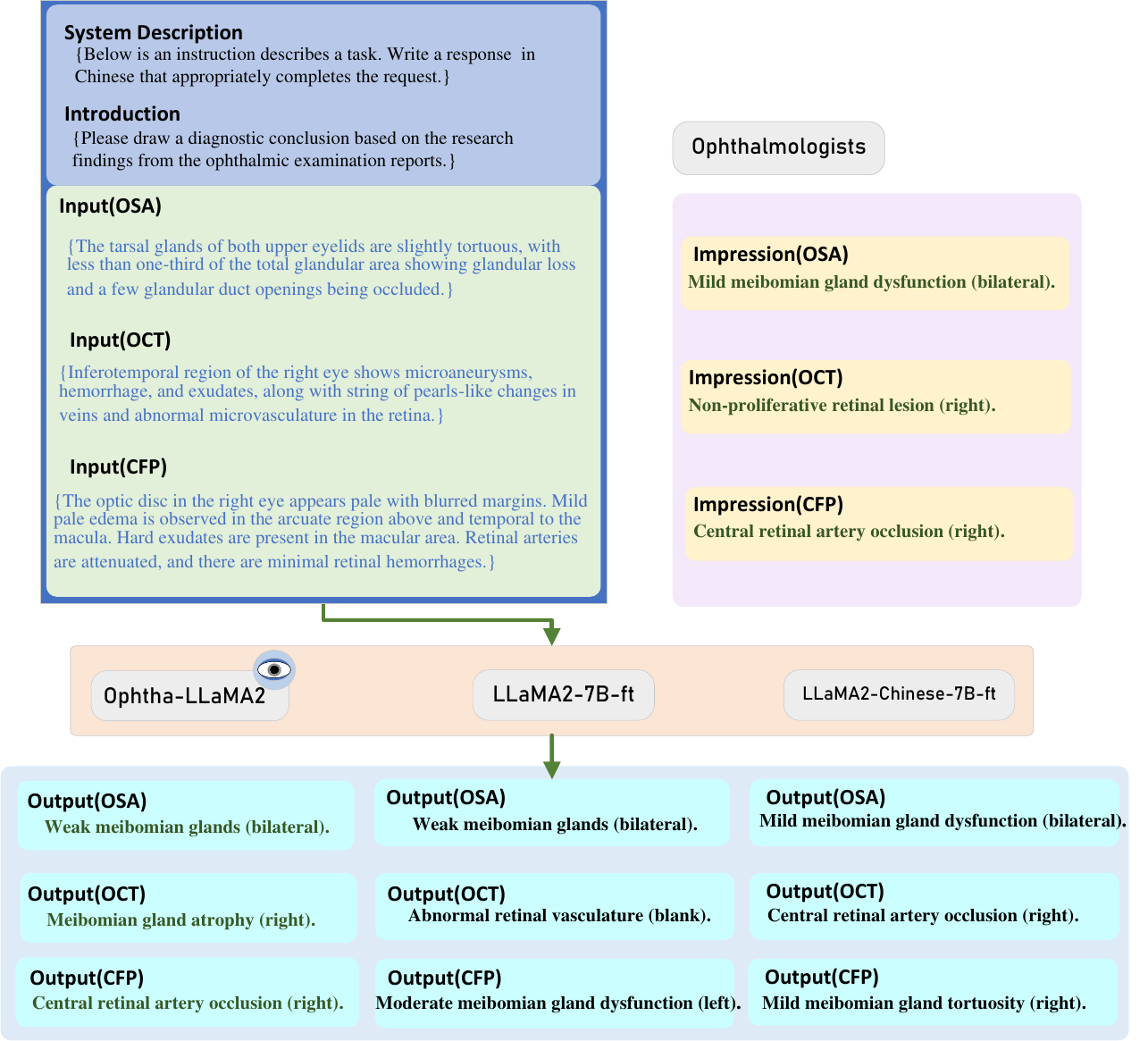}
    \caption{Results Visualization}
    \label{fig:Results}
\end{figure}

Based on the aforementioned engineering metrics and computational resource assessments, it can be concluded that Ophtha-LLaMA2 possesses advantages in terms of low energy consumption and high efficiency. Despite fine-tuning the model with a limited amount of ophthalmic data, it exhibited remarkable performance in engineering metrics, surpassing other models of the same scale.

The results of our evaluation indicate that Ophtha-LLaMA2 is capable of efficiently diagnosing eye diseases. Despite the limited amount of data used for fine-tuning, the model has demonstrated satisfactory performance in real-world applications. This further confirms the effectiveness of our fine-tuning approach and its ability to yield exceptional results within a short timeframe. Moreover, according to clinical evaluations, the model exhibits high accuracy and reliability in diagnosing various types of ophthalmic diseases. Figure \ref{fig:Results} illustrates an example result randomly selected from each of the three imaging modalities.

In conclusion, Ophtha-LLaMA2 offers significant advantages in terms of low energy consumption and high efficiency. Despite undergoing limited fine-tuning, the model has exhibited outstanding performance in engineering evaluations, surpassing other models of comparable scale. Overall, Ophtha-LLaMA2 provides a reliable and efficient solution for diagnosing eye diseases.

\section{Discussion}

\subsection{LLMs have tremendous potential in the field of ophthalmology}

LLMs, especial LLaMA2 have tremendous potential in the field of ophthalmology. Our study has shown that  with training on a small number of ophthalmic reports,  Ophtha-LLaMA2 is capable of accurately predicting and identifying various eye diseases and abnormalities. The impressive performance of LLMs such as LLaMA2 highlights their ability as powerful tools for assisting ophthalmologists in report diagnosis. In future research, we will attempt to combine multi-modal imaging for comprehensive analysis to improve the accuracy of anomaly localization and disease diagnosis.

\subsection{Applicability of the Rouge metric for evaluating medical diagnosis}

The Rouge metric is a commonly used evaluation metric in NLP tasks to measure the similarity between automatically generated and reference summaries. However, there may be some limitations in using the Rouge metrics to evaluate the accuracy and effectiveness of medical diagnosis.

First, the Rouge metric primarily focuses on text matching between summaries, while medical diagnosis relies on various data sources, including patient history information and pathological images, rather than simple text generation. Therefore, relying solely on text similarity may not fully capture the complexity of medical diagnosis.

Second, the accuracy and effectiveness of medical diagnosis require the consideration of a variety of factors, including patient history, clinical signs, and laboratory test results. The Rouge metrics cannot capture these crucial pieces of information, and hence may not accurately assess the quality of medical diagnoses.

Furthermore, the Rouge metrics typically rely on a single reference summary to evaluate the generated summaries. In medical diagnosis, there is often no single gold standard diagnosis. Different doctors may provide different diagnostic opinions based on their experiences and judgments. Therefore, incorporating multiple reference standards would allow better evaluation of the quality of medical diagnosis.

In conclusion, although the Rouge metrics are widely used in NLP, it may have several limitations in the evaluation of the reasonableness of medical diagnosis. It is important to consider various factors, including text similarity, clinical information, and expert knowledge, to comprehensively evaluate the accuracy and effectiveness of medical diagnosis. This is also one of the directions we hope to explore in the future, to find a more suitable quantitative metric that aligns with clinical diagnosis evaluation.

\section{Conclusion and Future Work}

In this study, we successfully demonstrated the potential of LLMs in the diagnosis of ophthalmic reports. By training on a large dataset of ophthalmic reports, Ophtha-LLaMA2 is able to accurately predict and identify various ophthalmic diseases and abnormalities. The results of this study indicate that LLMs hold great promise as a powerful tool for assisting ophthalmologists in report diagnosis.There are still many future directions that could be explored in this research. Potential avenues for future work may include:

\textbf{(1) Expand the Dataset}

Although the dataset used in this study is extensive, there may still be a lack of data on rare diseases or conditions. Future work could focus on collecting more comprehensive ophthalmic report data to further improve the accuracy and generalization capability of the model.

\textbf{(2) Incorporate Multimodal Information}

Currently, LLMs primarily rely on textual inputs and have not fully leveraged ophthalmic images or other modalities of information. Future research can explore how to integrate image and text information to train a large multimodal models (LMM) for providing more comprehensive diagnosis and analysis of ophthalmic reports. By integrating information from different modalities, such as ophthalmic images and textual descriptions, the accuracy of diagnosis can be enhanced, enabling more detailed diagnoses and personalized treatment recommendations. This will be one of the important directions for future research in the field of ophthalmology.

\textbf{(3) Integrate Clinical Data}

A large volume of clinical data can provide more background information for the model and help improve the accuracy of ophthalmic report diagnosis. Future work could focus on integrating clinical data, including medical history and physical examination results to enhance the model's overall capability.

\textbf{(4) Customized Diagnosis}

Each patient's ophthalmic report can exhibit individual differences, making customized diagnosis a crucial task. Future research can explore how to provide personalized ophthalmic report diagnosis based on a patient's specific circumstances, such as medical history, hospitalization records, and other relevant factors mentioned earlier. This will enhance the model's clinical applicability, providing doctors with diagnoses and treatment recommendations that are tailored to each patient's situation.

In conclusion, future work should concentrate on further enhancing the accuracy of LLMs, expanding their application scope, and increasing their integration with clinical practice. Through continuous improvement and innovation, we anticipate exciting advancements in the application of LLMs and LMMs in ophthalmic report diagnosis.

\section*{Acknowledgement}
This study is partly supported by National Natural Science Foundation of China (No. 62276050, 61976045, 82160195), Jiangxi Double-Thousand Plan High-Level Talent Project of Science and Technology Innovation (No. jxsq2023201036), and Key R \& D Program of Jiangxi Province (No. 20223BBH80014).

\bibliography{LLM_refs}

\begin{thebibliography}{10}

\bibitem{danilov2022length}
G~Danilov, K~Kotik, E~Shevchenko, D~Usachev, M~Shifrin, Y~Strunina, T~Tsukanova, T~Ishankulov, V~Lukshin, and A~Potapov.
\newblock Length of stay prediction in neurosurgery with russian gpt-3 language model compared to human expectations.
\newblock {\em Studies in Health Technology and Informatics}, 289:156--159, 2022.

\bibitem{wang2022leveraging}
Sophia~Y Wang, Justin Huang, Hannah Hwang, Wendeng Hu, Shiqi Tao, and Tina Hernandez-Boussard.
\newblock Leveraging weak supervision to perform named entity recognition in electronic health records progress notes to identify the ophthalmology exam.
\newblock {\em International Journal of Medical Informatics}, 167:104864, 2022.

\bibitem{holmes2023benchmarking}
Jason Holmes, Lian Zhang, Yuzhen Ding, Hongying Feng, Zhengliang Liu, Tianming Liu, William~W Wong, Sujay~A Vora, Jonathan~B Ashman, and Wei Liu.
\newblock Benchmarking a foundation llm on its ability to re-label structure names in accordance with the aapm tg-263 report.
\newblock {\em arXiv preprint arXiv:2310.03874}, 2023.

\bibitem{liu2023artificial}
Chenbin Liu, Zhengliang Liu, Jason Holmes, Lu~Zhang, Lian Zhang, Yuzhen Ding, Peng Shu, Zihao Wu, Haixing Dai, Yiwei Li, et~al.
\newblock Artificial general intelligence for radiation oncology.
\newblock {\em arXiv preprint arXiv:2309.02590}, 2023.

\bibitem{liu2023evaluating}
Zhengliang Liu, Tianyang Zhong, Yiwei Li, Yutong Zhang, Yi~Pan, Zihao Zhao, Peixin Dong, Chao Cao, Yuxiao Liu, Peng Shu, Yaonai Wei, Zihao Wu, Chong Ma, Jiaqi Wang, Sheng Wang, Mengyue Zhou, Zuowei Jiang, Chunlin Li, Jason Holmes, Shaochen Xu, Lu~Zhang, Haixing Dai, Kai Zhang, Lin Zhao, Yuanhao Chen, Xu~Liu, Peilong Wang, Pingkun Yan, Jun Liu, Bao Ge, Lichao Sun, Dajiang Zhu, Xiang Li, Wei Liu, Xiaoyan Cai, Xintao Hu, Xi~Jiang, Shu Zhang, Xin Zhang, Tuo Zhang, Shijie Zhao, Quanzheng Li, Hongtu Zhu, Dinggang Shen, and Tianming Liu.
\newblock Evaluating large language models for radiology natural language processing, 2023.

\bibitem{li2023artificial}
Xiang Li, Lu~Zhang, Zihao Wu, Zhengliang Liu, Lin Zhao, Yixuan Yuan, Jun Liu, Gang Li, Dajiang Zhu, Pingkuan Yan, et~al.
\newblock Artificial general intelligence for medical imaging.
\newblock {\em arXiv preprint arXiv:2306.05480}, 2023.

\bibitem{holmes2023evaluating}
J~Holmes, Z~Liu, L~Zhang, Y~Ding, TT~Sio, LA~McGee, JB~Ashman, X~Li, T~Liu, J~Shen, et~al.
\newblock Evaluating large language models on a highly-specialized topic.
\newblock {\em Radiation Oncology Physics}, 2023.

\bibitem{wu2023matching}
Xuansheng Wu, Xinyu He, Tianming Liu, Ninghao Liu, and Xiaoming Zhai.
\newblock Matching exemplar as next sentence prediction (mensp): Zero-shot prompt learning for automatic scoring in science education.
\newblock In {\em International Conference on Artificial Intelligence in Education}, pages 401--413. Springer, 2023.

\bibitem{latif2023artificial}
Ehsan Latif, Gengchen Mai, Matthew Nyaaba, Xuansheng Wu, Ninghao Liu, Guoyu Lu, Sheng Li, Tianming Liu, and Xiaoming Zhai.
\newblock Artificial general intelligence (agi) for education.
\newblock {\em arXiv preprint arXiv:2304.12479}, 2023.

\bibitem{liu2023context}
Zhengliang Liu, Xinyu He, Lei Liu, Tianming Liu, and Xiaoming Zhai.
\newblock Context matters: A strategy to pre-train language model for science education.
\newblock {\em arXiv preprint arXiv:2301.12031}, 2023.

\bibitem{ouyang2022training}
Long Ouyang, Jeffrey Wu, Xu~Jiang, Diogo Almeida, Carroll Wainwright, Pamela Mishkin, Chong Zhang, Sandhini Agarwal, Katarina Slama, Alex Ray, et~al.
\newblock Training language models to follow instructions with human feedback.
\newblock {\em Advances in Neural Information Processing Systems}, 35:27730--27744, 2022.

\bibitem{thirunavukarasu2023trialling}
Arun~James Thirunavukarasu, Refaat Hassan, Shathar Mahmood, Rohan Sanghera, Kara Barzangi, Mohanned El~Mukashfi, and Sachin Shah.
\newblock Trialling a large language model (chatgpt) in general practice with the applied knowledge test: observational study demonstrating opportunities and limitations in primary care.
\newblock {\em JMIR Medical Education}, 9(1):e46599, 2023.

\bibitem{thirunavukarasu2023large}
Arun~James Thirunavukarasu, Darren Shu~Jeng Ting, Kabilan Elangovan, Laura Gutierrez, Ting~Fang Tan, and Daniel Shu~Wei Ting.
\newblock Large language models in medicine.
\newblock {\em Nature medicine}, 29(8):1930--1940, 2023.

\bibitem{shuster2022blenderbot}
Kurt Shuster, Jing Xu, Mojtaba Komeili, Da~Ju, Eric~Michael Smith, Stephen Roller, Megan Ung, Moya Chen, Kushal Arora, Joshua Lane, et~al.
\newblock Blenderbot 3: a deployed conversational agent that continually learns to responsibly engage.
\newblock {\em arXiv preprint arXiv:2208.03188}, 2022.

\bibitem{glaese2022improving}
Amelia Glaese, Nat McAleese, Maja Tr{\k{e}}bacz, John Aslanides, Vlad Firoiu, Timo Ewalds, Maribeth Rauh, Laura Weidinger, Martin Chadwick, Phoebe Thacker, et~al.
\newblock Improving alignment of dialogue agents via targeted human judgements.
\newblock {\em arXiv preprint arXiv:2209.14375}, 2022.

\bibitem{moqurrab2021accurate}
Syed~Atif Moqurrab, Umair Ayub, Adeel Anjum, Sohail Asghar, and Gautam Srivastava.
\newblock An accurate deep learning model for clinical entity recognition from clinical notes.
\newblock {\em IEEE Journal of Biomedical and Health Informatics}, 25(10):3804--3811, 2021.

\bibitem{devlin2018bert}
Jacob Devlin, Ming-Wei Chang, Kenton Lee, and Kristina Toutanova.
\newblock Bert: Pre-training of deep bidirectional transformers for language understanding.
\newblock {\em arXiv preprint arXiv:1810.04805}, 2018.

\bibitem{chowdhery2022palm}
Aakanksha Chowdhery, Sharan Narang, Jacob Devlin, Maarten Bosma, Gaurav Mishra, Adam Roberts, Paul Barham, Hyung~Won Chung, Charles Sutton, Sebastian Gehrmann, et~al.
\newblock Palm: Scaling language modeling with pathways.
\newblock {\em arXiv preprint arXiv:2204.02311}, 2022.

\bibitem{floridi2020gpt}
Luciano Floridi and Massimo Chiriatti.
\newblock Gpt-3: Its nature, scope, limits, and consequences.
\newblock {\em Minds and Machines}, 30:681--694, 2020.

\bibitem{openai2023gpt4}
OpenAI.
\newblock Gpt-4 technical report, 2023.

\bibitem{yang2023dawn}
Zhengyuan Yang, Linjie Li, Kevin Lin, Jianfeng Wang, Chung-Ching Lin, Zicheng Liu, and Lijuan Wang.
\newblock The dawn of lmms: Preliminary explorations with gpt-4v(ision), 2023.

\bibitem{openai2023gpt4v}
OpenAI.
\newblock Gpt-4v(ision) system card, 2023.

\bibitem{nori2023capabilities}
Harsha Nori, Nicholas King, Scott~Mayer McKinney, Dean Carignan, and Eric Horvitz.
\newblock Capabilities of gpt-4 on medical challenge problems, 2023.

\bibitem{harskamp2023performance}
Ralf~E Harskamp and Lukas De~Clercq.
\newblock Performance of chatgpt as an ai-assisted decision support tool in medicine: a proof-of-concept study for interpreting symptoms and management of common cardiac conditions (amstelheart-2).
\newblock {\em medRxiv}, pages 2023--03, 2023.

\bibitem{rao2023evaluating}
Arya Rao, John Kim, Meghana Kamineni, Michael Pang, Winston Lie, Keith~J Dreyer, and Marc~D Succi.
\newblock Evaluating gpt as an adjunct for radiologic decision making: Gpt-4 versus gpt-3.5 in a breast imaging pilot.
\newblock {\em Journal of the American College of Radiology}, 2023.

\bibitem{liu2023summary}
Yiheng Liu, Tianle Han, Siyuan Ma, Jiayue Zhang, Yuanyuan Yang, Jiaming Tian, Hao He, Antong Li, Mengshen He, Zhengliang Liu, et~al.
\newblock Summary of chatgpt/gpt-4 research and perspective towards the future of large language models.
\newblock {\em arXiv preprint arXiv:2304.01852}, 2023.

\bibitem{nov2023putting}
Oded Nov, Nina Singh, and Devin~M Mann.
\newblock Putting chatgpt's medical advice to the (turing) test.
\newblock {\em medRxiv}, pages 2023--01, 2023.

\bibitem{shi2023mededit}
Yucheng Shi, Shaochen Xu, Zhengliang Liu, Tianming Liu, Xiang Li, and Ninghao Liu.
\newblock Mededit: Model editing for medical question answering with external knowledge bases.
\newblock {\em arXiv preprint arXiv:2309.16035}, 2023.

\bibitem{wang2023prompt}
Jiaqi Wang, Enze Shi, Sigang Yu, Zihao Wu, Chong Ma, Haixing Dai, Qiushi Yang, Yanqing Kang, Jinru Wu, Huawen Hu, et~al.
\newblock Prompt engineering for healthcare: Methodologies and applications.
\newblock {\em arXiv preprint arXiv:2304.14670}, 2023.

\bibitem{zhang2023biomedgpt}
Kai Zhang, Jun Yu, Zhiling Yan, Yixin Liu, Eashan Adhikarla, Sunyang Fu, Xun Chen, Chen Chen, Yuyin Zhou, Xiang Li, et~al.
\newblock Biomedgpt: A unified and generalist biomedical generative pre-trained transformer for vision, language, and multimodal tasks.
\newblock {\em arXiv preprint arXiv:2305.17100}, 2023.

\bibitem{liu2023pharmacygpt}
Zhengliang Liu, Zihao Wu, Mengxuan Hu, Bokai Zhao, Lin Zhao, Tianyi Zhang, Haixing Dai, Xianyan Chen, Ye~Shen, Sheng Li, et~al.
\newblock Pharmacygpt: The ai pharmacist.
\newblock {\em arXiv preprint arXiv:2307.10432}, 2023.

\bibitem{cai2023exploring}
Hongmin Cai, Xiaoke Huang, Zhengliang Liu, Wenxiong Liao, Haixing Dai, Zihao Wu, Dajiang Zhu, Hui Ren, Quanzheng Li, Tianming Liu, et~al.
\newblock Exploring multimodal approaches for alzheimer's disease detection using patient speech transcript and audio data.
\newblock {\em arXiv preprint arXiv:2307.02514}, 2023.

\bibitem{zhong2023chatabl}
Tianyang Zhong, Yaonai Wei, Li~Yang, Zihao Wu, Zhengliang Liu, Xiaozheng Wei, Wenjun Li, Junjie Yao, Chong Ma, Xiang Li, et~al.
\newblock Chatabl: Abductive learning via natural language interaction with chatgpt.
\newblock {\em arXiv preprint arXiv:2304.11107}, 2023.

\bibitem{wang2023review}
Jiaqi Wang, Zhengliang Liu, Lin Zhao, Zihao Wu, Chong Ma, Sigang Yu, Haixing Dai, Qiushi Yang, Yiheng Liu, Songyao Zhang, et~al.
\newblock Review of large vision models and visual prompt engineering.
\newblock {\em arXiv preprint arXiv:2307.00855}, 2023.

\bibitem{dai2023ad}
Haixing Dai, Yiwei Li, Zhengliang Liu, Lin Zhao, Zihao Wu, Suhang Song, Ye~Shen, Dajiang Zhu, Xiang Li, Sheng Li, et~al.
\newblock Ad-autogpt: An autonomous gpt for alzheimer's disease infodemiology.
\newblock {\em arXiv preprint arXiv:2306.10095}, 2023.

\bibitem{guan2023cohortgpt}
Zihan Guan, Zihao Wu, Zhengliang Liu, Dufan Wu, Hui Ren, Quanzheng Li, Xiang Li, and Ninghao Liu.
\newblock Cohortgpt: An enhanced gpt for participant recruitment in clinical study.
\newblock {\em arXiv preprint arXiv:2307.11346}, 2023.

\bibitem{zhong2023chatradio}
Tianyang Zhong, Wei Zhao, Yutong Zhang, Yi~Pan, Peixin Dong, Zuowei Jiang, Xiaoyan Kui, Youlan Shang, Li~Yang, Yaonai Wei, et~al.
\newblock Chatradio-valuer: A chat large language model for generalizable radiology report generation based on multi-institution and multi-system data.
\newblock {\em arXiv preprint arXiv:2310.05242}, 2023.

\bibitem{liu2023deid}
Zhengliang Liu, Xiaowei Yu, Lu~Zhang, Zihao Wu, Chao Cao, Haixing Dai, Lin Zhao, Wei Liu, Dinggang Shen, Quanzheng Li, et~al.
\newblock Deid-gpt: Zero-shot medical text de-identification by gpt-4.
\newblock {\em arXiv preprint arXiv:2303.11032}, 2023.

\bibitem{dai2023auggpt}
Haixing Dai, Zhengliang Liu, Wenxiong Liao, Xiaoke Huang, Yihan Cao, Zihao Wu, Lin Zhao, Shaochen Xu, Wei Liu, Ninghao Liu, Sheng Li, Dajiang Zhu, Hongmin Cai, Lichao Sun, Quanzheng Li, Dinggang Shen, Tianming Liu, and Xiang Li.
\newblock Auggpt: Leveraging chatgpt for text data augmentation, 2023.

\bibitem{xu2022altered}
Yu-Ling Xu, Xiao-Yu Wang, Jun Chen, Min Kang, Yi-Xin Wang, Li-Juan Zhang, Hui-Ye Shu, Xu-Lin Liao, Jie Zou, Hong Wei, et~al.
\newblock Altered spontaneous brain activity patterns of meibomian gland dysfunction in severely obese population measured using the fractional amplitude of low-frequency fluctuations.
\newblock {\em Frontiers in Psychiatry}, 13:914039, 2022.

\bibitem{liao2019altered}
Xu-Lin Liao, Qing Yuan, Wen-Qing Shi, Biao Li, Ting Su, Qi~Lin, You-Lan Min, Pei-Wen Zhu, Lei Ye, and Yi~Shao.
\newblock Altered brain activity in patients with diabetic retinopathy using regional homogeneity: a resting-state fmri study.
\newblock {\em Endocrine Practice}, 25(4):320--327, 2019.

\bibitem{shao2018visualization}
Yi~Shao, Hong Jiang, Yantao Wei, Yingying Shi, Ce~Shi, Clinton~B Wright, Xiaoyan Sun, Elizabeth~A Vanner, Anny~D Rodriguez, Byron~L Lam, et~al.
\newblock Visualization of focal thinning of the ganglion cell--inner plexiform layer in patients with mild cognitive impairment and alzheimer’s disease.
\newblock {\em Journal of Alzheimer's disease}, 64(4):1261--1273, 2018.

\bibitem{ye2018retinal}
Lei Ye, Shuang-Shuang Zhou, Wen-Long Yang, Jing Bao, Nan Jiang, You-Lan Min, Qing Yuan, Gang Tan, Mei Shen, and Yi~Shao.
\newblock Retinal microvasculature alteration in active thyroid-associated ophthalmopathy.
\newblock {\em Endocrine Practice}, 24(7):658--667, 2018.

\bibitem{touvron2023llama}
Hugo Touvron, Louis Martin, Kevin Stone, Peter Albert, Amjad Almahairi, Yasmine Babaei, Nikolay Bashlykov, Soumya Batra, Prajjwal Bhargava, Shruti Bhosale, et~al.
\newblock Llama 2: Open foundation and fine-tuned chat models.
\newblock {\em arXiv preprint arXiv:2307.09288}, 2023.

\bibitem{hu2021lora}
Edward~J Hu, Yelong Shen, Phillip Wallis, Zeyuan Allen-Zhu, Yuanzhi Li, Shean Wang, Lu~Wang, and Weizhu Chen.
\newblock Lora: Low-rank adaptation of large language models.
\newblock {\em arXiv preprint arXiv:2106.09685}, 2021.

\bibitem{lin2004rouge}
Chin-Yew Lin.
\newblock Rouge: A package for automatic evaluation of summaries.
\newblock In {\em Text summarization branches out}, pages 74--81, 2004.

\end{thebibliography}
\bibliographystyle{unsrt}

\end{document}